\documentclass[3p,,preprint,12pt,onecolumn]{elsarticle}

\usepackage{lineno,hyperref}
\modulolinenumbers[5]
\usepackage{units}
\usepackage[usenames, dvipsnames]{color}
\usepackage{booktabs}
\usepackage{siunitx}
\usepackage{multirow}

\journal{Journal of \LaTeX\ Templates}

\long\def\twocolumn[#1]{#1}
\makeatletter\@twocolumnfalse \makeatother

\usepackage{tabulary,xcolor}
\usepackage{amsfonts,amsmath,amssymb}
\usepackage[T1]{fontenc}
\makeatletter
\let\save@ps@pprintTitle\ps@pprintTitle
\def\ps@pprintTitle{\save@ps@pprintTitle\gdef\@oddfoot{\footnotesize\itshape \null\hfill\today}}
\def\hlinewd#1{%
  \noalign{\ifnum0=`}\fi\hrule \@height #1%
  \futurelet\reserved@a\@xhline}

\AtBeginDocument{\ifNAT@numbers \biboptions{sort&compress}\fi}
\makeatother
\usepackage{pgfplots}    
\usepackage{ifluatex}
\ifluatex
\usepackage{fontspec}
\defaultfontfeatures{Ligatures=TeX}
\usepackage[]{unicode-math}
\unimathsetup{math-style=TeX}
\else 
\usepackage[utf8]{inputenc}
\fi 
\ifluatex\else\usepackage{stmaryrd}\fi
  
\usepackage{url,multirow,morefloats,floatflt,cancel,tfrupee}
\makeatletter

\def\@print@issn{\@ifundefined{iet@jnl@issn}{}{\iet@jnl@issn}}

\AtBeginDocument{\@ifpackageloaded{textcomp}{}{\usepackage{textcomp}}}
\makeatother
\usepackage{graphicx}
\usepackage{subfigure}
\usepackage{longtable}
\usepackage[percent]{overpic}
\usepackage{colortbl}
\usepackage{xcolor}
\usepackage{pifont}
\usepackage{extarrows}
\usepackage[nointegrals]{wasysym}
\makeatletter

\def\mcWidth#1{\csname TY@F#1\endcsname+\tabcolsep}

\def\cAlignHack{\rightskip\@flushglue\leftskip\@flushglue\parindent\z@\parfillskip\z@skip}
\def\rAlignHack{\rightskip\z@skip\leftskip\@flushglue \parindent\z@\parfillskip\z@skip}

\if@twocolumn\usepackage{dblfloatfix}\fi 
\AtBeginDocument{
\expandafter\ifx\csname eqalign\endcsname\relax
\def\eqalign#1{\null\vcenter{\def\\{\cr}\openup\jot\m@th
  \ialign{\strut$\displaystyle{##}$\hfil&$\displaystyle{{}##}$\hfil
      \crcr#1\crcr}}\,}
\fi
}

\let\lt=<
\let\gt=>
\def\processVert{\ifmmode|\else\textbar\fi}

\@ifundefined{subparagraph}{
\def\subparagraph{\@startsection{paragraph}{5}{2\parindent}{0ex plus 0.1ex minus 0.1ex}%
{0ex}{\normalfont\small\itshape}}%
}{}

\newcommand\role[1]{\unskip}
\newcommand\aucollab[1]{\unskip}
  
\@ifundefined{tsGraphicsScaleX}{\gdef\tsGraphicsScaleX{1}}{}
\@ifundefined{tsGraphicsScaleY}{\gdef\tsGraphicsScaleY{.9}}{}
\def\checkGraphicsWidth{\ifdim\Gin@nat@width>\linewidth
	\tsGraphicsScaleX\linewidth\else\Gin@nat@width\fi}

\def\checkGraphicsHeight{\ifdim\Gin@nat@height>.9\textheight
	\tsGraphicsScaleY\textheight\else\Gin@nat@height\fi}

\def\fixFloatSize#1{}
\let\ts@includegraphics\includegraphics

\def\inlinegraphic[#1]#2{{\edef\@tempa{#1}\edef\baseline@shift{\ifx\@tempa\@empty0\else#1\fi}\edef\tempZ{\the\numexpr(\numexpr(\baseline@shift*\f@size/100))}\protect\raisebox{\tempZ pt}{\ts@includegraphics{#2}}}}
\renewcommand{\includegraphics}[1]{\ts@includegraphics[width=\checkGraphicsWidth]{#1}}
\AtBeginDocument{\def\includegraphics{\@ifnextchar[{\ts@includegraphics}{\ts@includegraphics[width=\checkGraphicsWidth,height=\checkGraphicsHeight,keepaspectratio]}}}

\def\URL#1#2{\@ifundefined{href}{#2}{\href{#1}{#2}}}

\def\UrlOrds{\do\*\do\-\do\~\do\'\do\"\do\-}%
\g@addto@macro{\UrlBreaks}{\UrlOrds}
\makeatother

\begin{document}

\begin{frontmatter}
	
\title{A 2D laser rangefinder scans dataset of standard EUR pallets}
    
\author[inria]{Ihab S. Mohamed\corref{contrib-inria}}
\ead{ihab.mohamed@inria.fr}\cortext[contrib-inria]{Corresponding author.}
\author[Dibris]{Alessio Capitanelli}
\ead{alessio.capitanelli@dibris.unige.it}
\author[Dibris]{Fulvio Mastrogiovanni}
\ead{fulvio.mastrogiovanni@unige.it}
\author[Dibris]{Stefano Rovetta}
\ead{stefano.rovetta@unige.it}
\author[Dibris]{Renato Zaccaria}
\ead{renato.zaccaria@unige.it}
   
\address[inria]{INRIA Sophia Antipolis - Méditerranée\unskip, 
     Université Côte d'Azur\unskip, France}
           
\address[Dibris]{Department of Informatics, Bioengineering, Robotics and Systems Engineering (DIBRIS)\unskip, 
     University of Genoa\unskip, Italy}

\begin{abstract}
In the past few years, the technology of automated guided vehicles (AGVs) has notably advanced. In particular, in the context of factory and warehouse automation, different approaches have been presented for detecting and localizing pallets inside warehouses and shop-floor environments. In a related research paper \cite{mohamed2018detection}, we show that an AGVs can detect, localize, and track pallets using machine learning techniques based only on the data of an on-board 2D laser rangefinder. Such sensor is very common in industrial scenarios due to its simplicity and robustness, but it can only provide a limited amount of data. Therefore, it has been neglected in the past in favor of more complex solutions. In this paper, we release to the community the data we collected in \citep{mohamed2018detection}
for further research activities in the field of pallet localization and tracking. The dataset comprises a collection of 565 2D scans from real-world environments, which are divided into 340 samples where pallets are present, and 225 samples where they are not. The data have been manually labelled and are provided in different formats.

\end{abstract} 
\begin{keyword} 
    2D Laser Rangefinder\sep Object Detection\sep Robotics\sep Automated Guided Vehicle
\end{keyword}
	
\end{frontmatter}


\section*{Specifications Table}  
\noindent
\begin{tabular}{|l|l|}
\hline\hline
\textbf{Subject area} & \textbf{Engineering}\\
\hline\hline
More specific subject area & Robotics, Object Detection, Automated Guided Vehicle\\
\hline
Type of data & Raw depth data provided by the 2D range sensor\\
			 & Processed 2D bitmap-like image representation of raw data\\
\hline
How data was acquired & 2D laser rangefinder (SICK S3000 Pro CMS)\\
\hline
Data format & Files in text format \textit{.txt}\\
			& 2D images in \textit{.jpg $\&$ .png} ($681\times533$ \& $250\times250$ \textit{pixels})\\
			& MAT-files in MATLAB format \textit{.mat}\\
\hline
Experimental factors & 2D depth data processed offline and converted into 2D images.\\
                     & Images have been manually tagged whether they include a pallet\\
                     & or not, and eventually paired with the respective region of interest.\\
\hline
Experimental features & Raw data have been acquired by moving a 2D laser scanner\\
                      & in a realistic reproduction of a factory workshop, featuring\\
                      & pallets, people, robots and other equipment. \\
\hline
Data source location & EMARO Lab, Department of Informatics, Bioengineering, Robotics\\
                     & and Systems Engineering, University of Genoa, Genoa, Italy\\
                     & (44.402241, 8.960811)\\
\hline
Data accessibility & Dataset and codes are archived in a GitHub repository at: \\
				   & \url{https://github.com/EmaroLab/PDT}\\
\hline
Related research article & "Detection, localisation and tracking of pallets using\\
                         & learning techniques and 2D range data" \cite{mohamed2018detection}\\
\hline
\end{tabular}
\section*{Value of the Data}
\begin{itemize}
\item The 2D Laser Rangefinder dataset allows to develop novel techiques for pallet detection, localization and tracking. 
\item The 2D Laser Rangefinder dataset can be used as banchmark to compare the accuracy of different pallet detection, localization and tracking algorithms.
\item The 2D Laser Rangefinder dataset allows to improve Automated Guidance Vehicles in industrial workshop environments.
\item The 2D Laser Rangefinder dataset can be used to simulate the 2D range sensor data of a mobile robot moving in an industrial workshop environment.
\item To our knowledge, this is the first dataset for pallet localization and tracking using only 2D Laser Rangefinder data, as opposed to previous datasets aimed at generic AGV and/or more complex sensors \cite{geiger2013vision, maddern20171, jeong2018complex, rennie2016dataset}.
\end{itemize}

\section{Data}

In this article, we present a dataset of 2D range data obtained from a laser scanner moving inside an industrial workshop environment, where EUR standard pallets (see Fig. \ref{fig:pallet}), people, robots and other equipment are present. Each frame of the sensor trajectory corresponds to:
(i) a 2D range scan (see Table \ref{TAB:label}) obtained from a SICK S3000 Pro CMS laser rangefinder (see Fig. \ref{fig:sick3000});
(ii) a 2D image obtained by processing the 2D range scan (see Fig. \ref{fig:pallet_examples});
(iii) a tag attached by a human, indicating whether the scan includes a pallet or not;
and (iv), the region of interest of the pallet in the image (if any), also defined by a human.

The laser rangefinder has a resolution of \unit[0.25]{deg} and a maximum field of view of \unit[190]{deg}, leading to scans made of 761 ranges. It operates at \unit[16]{Hz} frequency, and the scans are averaged every $4$ frames during the static data acquisition phase in order to reduce noise. There are a total of $565$ scans, $340$ of which contains a pallet, while the remaining $225$ do not. The corresponding  2D images are obtained by converting the range data from polar to cartesian coordinates and resizing them to \unit[$250\times250$]{px}. Also, images containing a pallet come with a pallet Region Of Interest (ROI), defined by its upper-left and lower-right vertices. Finally, an additional set of $4$ continuous trajectories' raw range data is also made available, to allow online testing.

\section{Experimental design, materials, and methods}
\label{sec:Experimental_Setup}

\subsection{Equipment and Software}
In our experiment, the data have been acquired using a commercial 2D laser rangefinder from SICK, in particular the model S3000 Pro CMS\footnote{\url{https://www.sick.com/ag/en/s3000-professional-cms-sensor-head-with-io-module/s30a-6011db/p/p31284}} pictured in Fig. \ref{fig:sick3000}. The sensor has a maximum range of \unit[49]{m} (\unit[20]{m} at $20\%$ reflectivity), a resolution of \unit[0.25]{deg}, a \unit[16]{Hz} refresh frequency, and an empirical error of \unit[30]{mm}. The maximum field of view of the rangefinder is \unit[190]{deg}, which is sufficient for the detection of objects in front of an eventual AGV. The sensor generates an array of $761$ distances in polar coordinates, i.e., each value in the array correspond to the distance to the closest object for every angle in \unit[0.25]{deg} increments.

The choice of this sensor was due to its widespread adoption in industrial mobile robotics, where it is mostly employed for safety applications and is appreciated for its robustness and precision. It belongs to the class of sensors based on the Time-of-Flight (TOF) principle, i.e., sensors which measure the delay between the emission of a signal and the moment it hits back a receiver in order to estimate the distance to a surface. This category of systems involves sensing devices such as Laser Measurement Systems (LMS) or LIDARs, radars, TOF cameras, and sonar sensors, and they emit either light or sound waves in the environment. Knowing the speed with which the signal propagates and using precise circuitry to measure the exact TOF, the distance can be estimated with high precision.

The laser rangefinder is then connected to a PC through a RS422-USB converter, which has a transmission rate of \unit[500]{kBaud}. The PC used to acquire the data is equipped with an Intel\textsuperscript{\textregistered} Core i5-4210U \unit[1.70]{GHz} CPU and \unit[6]{GB} of RAM, and runs Ubuntu 16.04 64 bit. 

On the software side, real-world data is acquired online using an \emph{ad hoc} software\footnote{\url{https://github.com/RobotnikAutomation/s3000_laser}} running in the Robot Operating System framework\footnote{\url{http://www.ros.org/about-ros/}} (ROS). Offline processing (i.e., conversion to 2D images and manual definition of the regions of interest) has been perfomed in MATLAB. The scripts employed to that purpose and the resulting \textit{.mat} files are also provided as part of this dataset.

\subsection{Environment}
\begin{figure}[ht!]
\centering
\includegraphics[width=0.6\columnwidth]{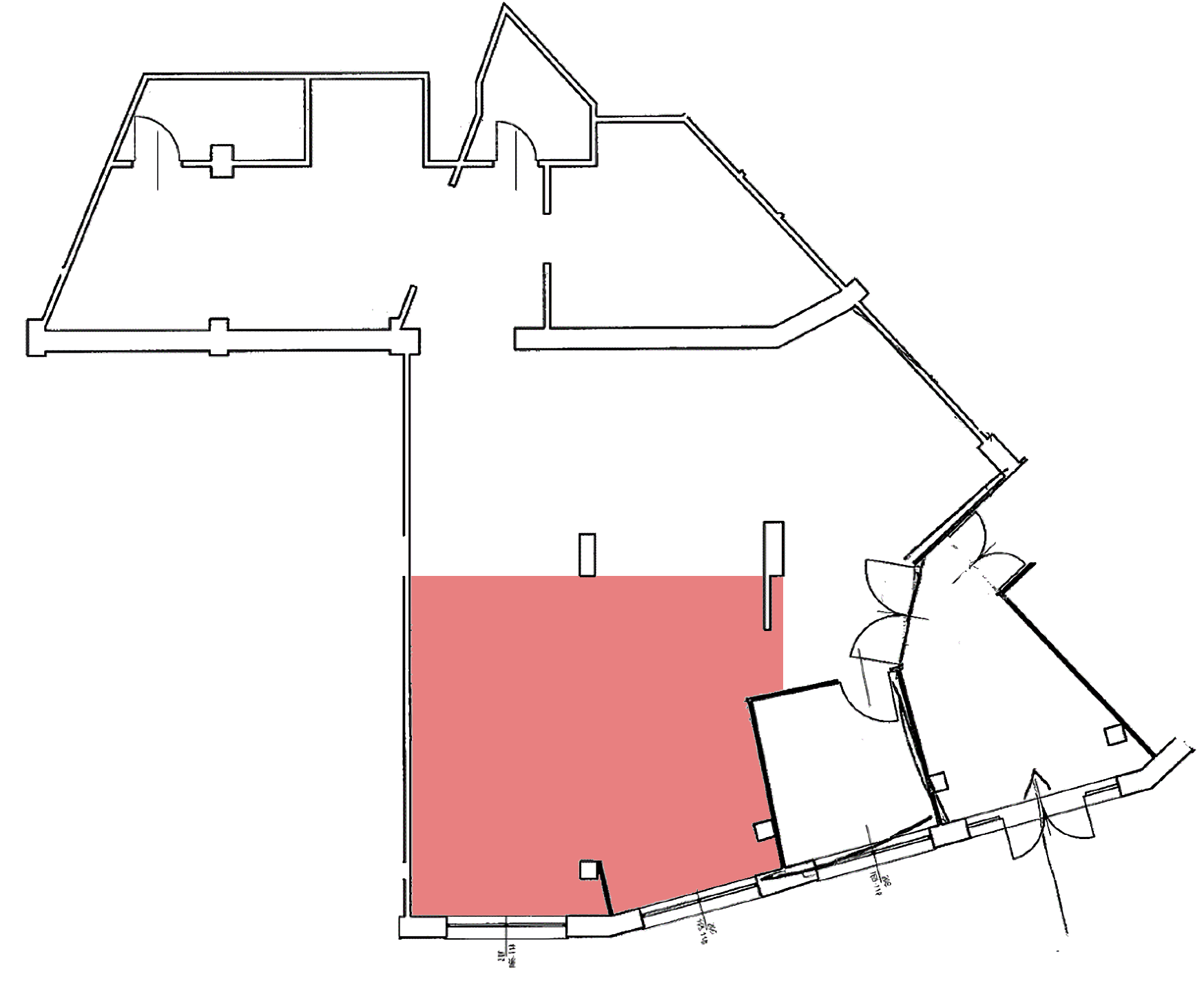}
\caption{A planimetry of the indoor environment where the experiment took place. The 2D laser rangefinder has been moved along several trajectories inside the read area, measuring \unit[40]{m$^2$}. The rest of the environmentis is still visible in several frames. In the whole environments several pieces of furniture and equipment, pallets, robots as well as people were present.}
\label{fig:labplan}
\end{figure}

\begin{figure}[ht!]
\centering
\subfigure{
\includegraphics[width=0.45\columnwidth]{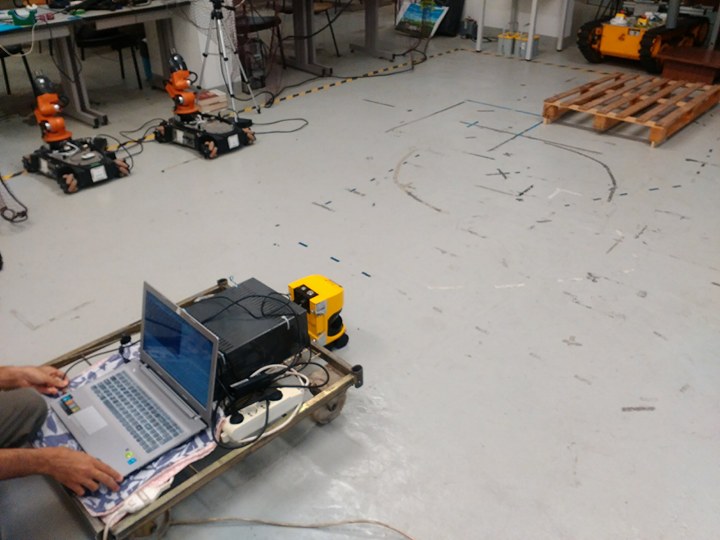}
\label{fig:test1}}
\subfigure{
\includegraphics[width=0.45\columnwidth]{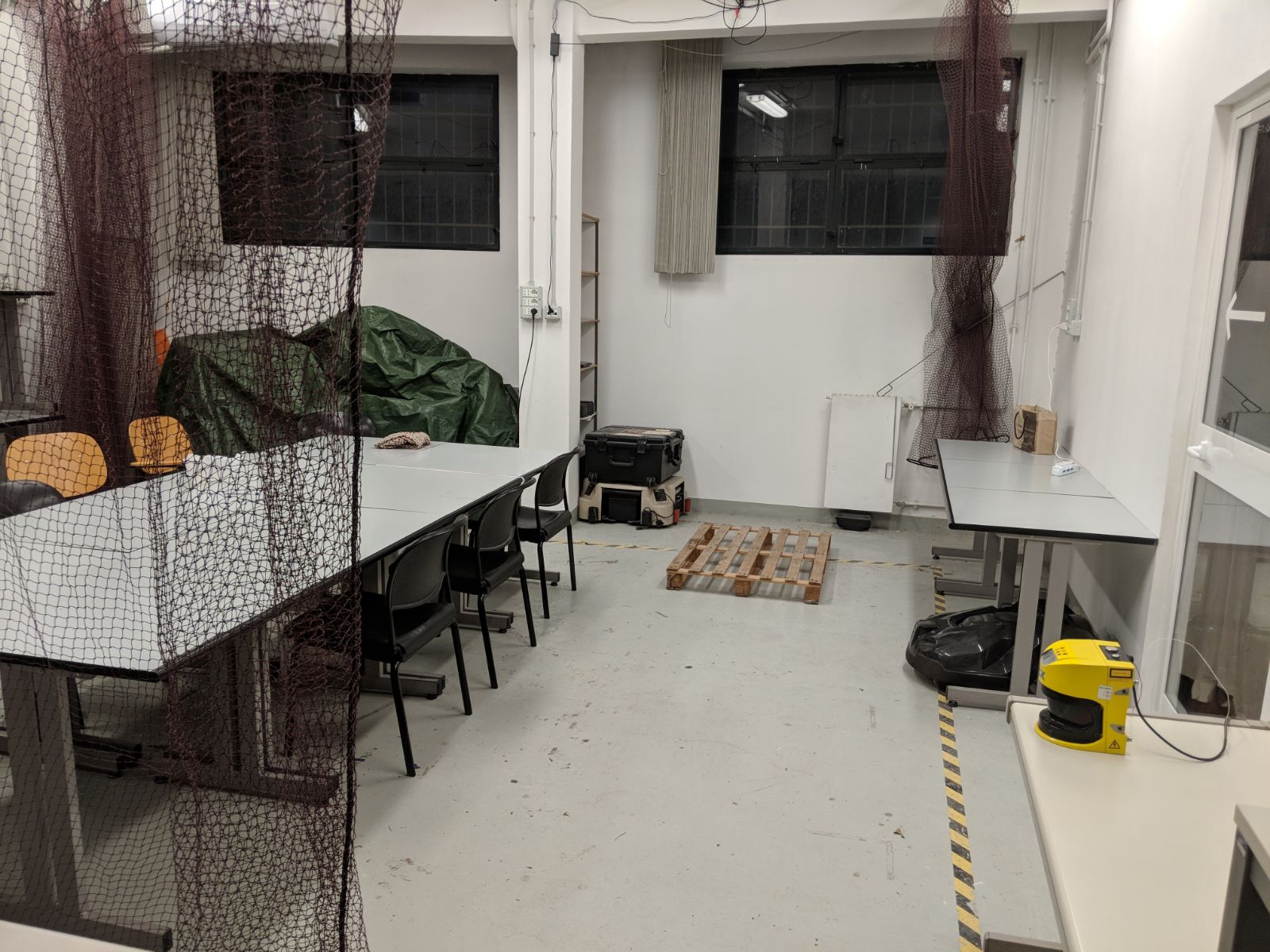}
\label{fig:test2}}
\subfigure{
\includegraphics[width=0.45\columnwidth]{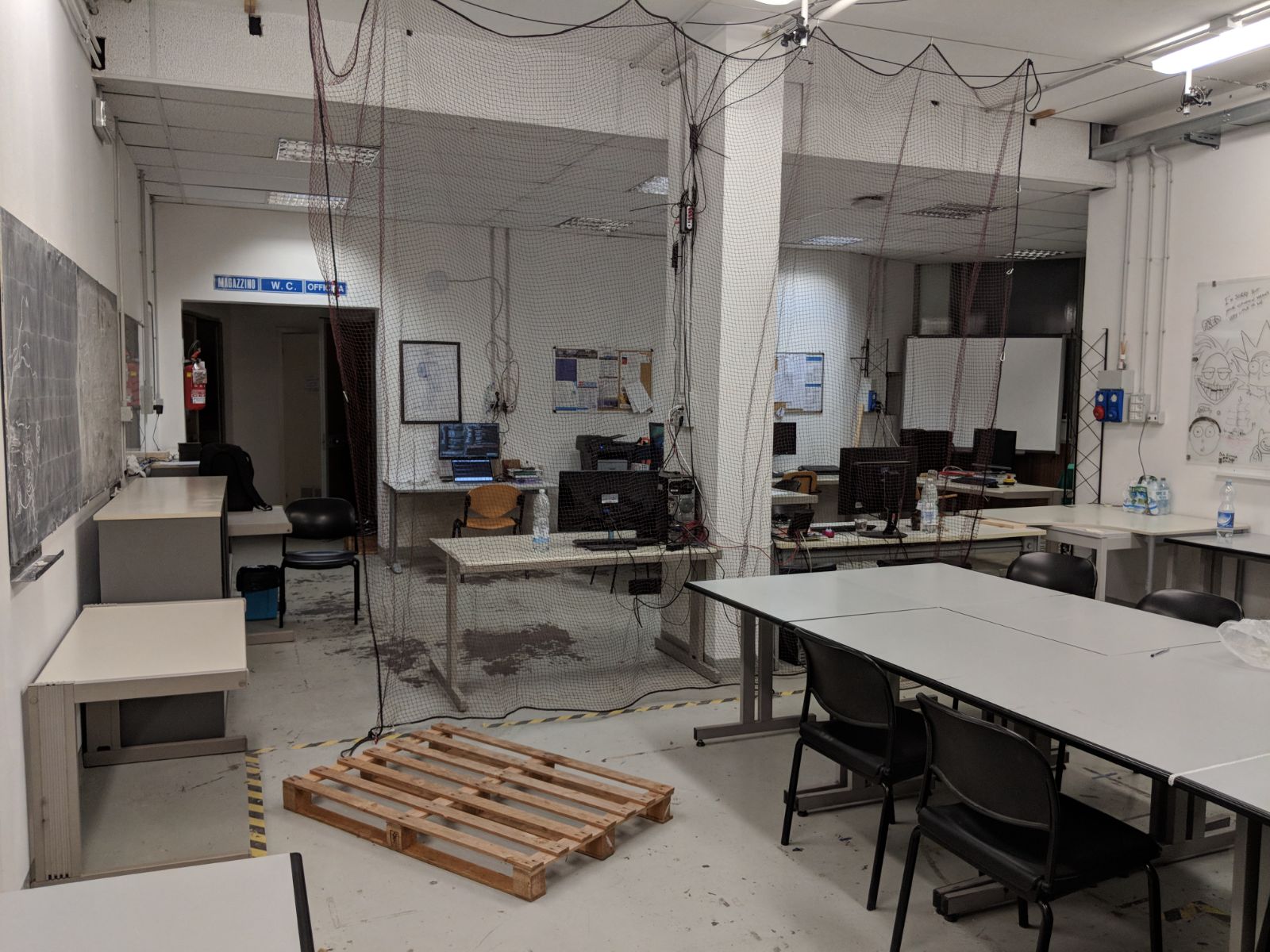}
\label{fig:test3}}
\subfigure{
\includegraphics[width=0.45\columnwidth]{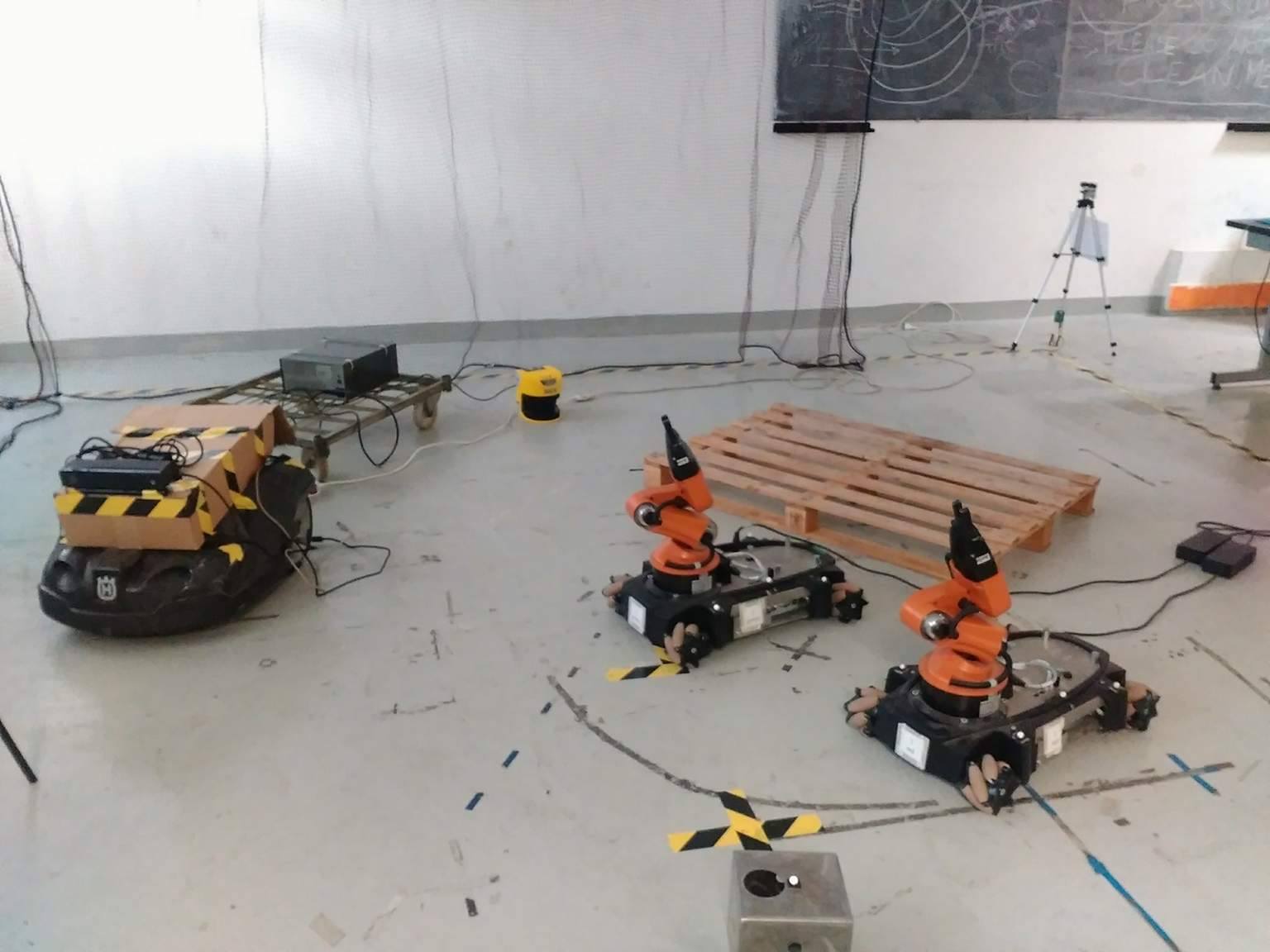}
\label{fig:test4}}
\caption{Snaphots of the test environment in different configurations. In the images a number of other objects appear beyond pallets, such as other robots, equipment and furniture.} 
\label{fig:testArea}
\end{figure}

We performed our experiments for data acquisition in the indoor environment represented in Figs. \ref{fig:labplan}-\ref{fig:testArea}, with the sensor moving in the \unit[40]{m$^2$} area highlighted in Fig. \ref{fig:labplan}. Such environmennt has been fitted to reproduce a typycal industrial workshop, featuring industrial pallets, furniture, robots and equipment (e.g., a conveyor belt). People were also included in the scene and allowed to move during data acquisition, which lead to temporary occlusions of the objects in the environment. Between acquisition sessions, the position of several objects was modified to better simulate a dynamic environment. The 2D laser rangefinder was positioned close to the floor, in a way that was both realistic with real world mounting position and able to perceive a pallet laying directly on the ground.

\begin{figure}[ht!]
\centering
\subfigure{
\includegraphics[width=0.55\columnwidth]{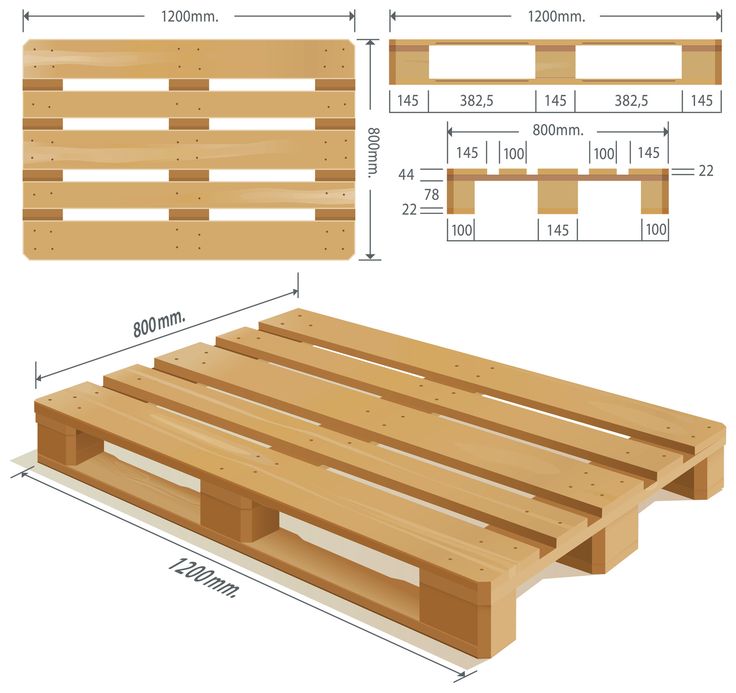}
\put (-160,250) {\textit{EUR-pallet}}
\label{fig:pallet}}
\subfigure{\raisebox{0.79in}
{\includegraphics[width=0.35\columnwidth]{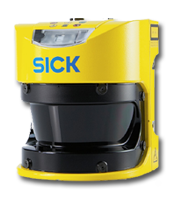}
\put (-150,190) {\textit{S3000 Professional CMS}}}
\label{fig:sick3000}}
\caption{The equipment that has been used to acquire the raw 2D range data: on the left hand side, the geometric characteristics of standard European pallet are shown, whilst on the right hand side the S3000 Professional laser scanner (Type: S30A-6011DB) is represented.}
\label{fig:setup}
\end{figure}

Concerning the type of pallet, we focused on the EUR-pallet standard depicted in Fig.\ref{fig:pallet}, which is the European pallet format specified by the European Pallet Association (EPAL)\footnote{\url{https://en.wikipedia.org/wiki/EUR-pallet}}. The size of EUR-pallets is \unit[1200]{mm}$\times$\unit[800]{mm} with a height of \unit[144]{mm}. Moreover, we defined as operating face of the pallet the one of narrower width. On that face there are two slots, each \unit[227.5]{mm} wide.

\subsection{Experiments}
In our experiments, the sensor was moved around the environment. Sensor frames differ from each other by the position and orientation of the pallet with respect of the sensor, but also due to the dynamic nature of the environment, as described in the previous section. In particular, it is possible that the pallet is heavily occluded and only few points belonging to it are visible in the frame. 

The acquired raw range data $R_i$ at any time instant $i$ represent the array of measured distances from the rangefinder to surrounding objects in the environment in the direction given by the angle $\phi_j$. More formally:
\begin{equation}\label{eq:1}
R_i = \lbrace r_0, \ldots,r_j, \ldots, r_{M}\rbrace,
\end{equation}
where $M$ is the maximum number of range points acquired per frame, which is related to the sensor's field of view and angular resolution. In our case, $M=761$, as the two values are \unit[190]{deg} and \unit[0.25]{deg} respectively. Keep in mind that the sensor employed runs at \unit[16]{Hz}, which would rapidly lead to a unmanageable amount of data, especially considering the manual labelling steps ahead. For this reason, we decided to effectively reduce the operating frequency to \unit[4]{Hz} in the static data acquisition phase, thus every $R_i$ is actually the result of the average of $4$ raw consecutive frames from the sensor. This also helps reducing noise on the data. An example of such process as well and the structure of the raw range data are reported in Table \ref{TAB:label}.

In our experiments, we are focusing on the detection of pallets in the environment, hence, the set $R$ of all raw range data readings $R_i$, consisting of $565$ 2D range scans, has been manually divided into two classes:

\begin{enumerate}
\item \textit{Pallet} class represents the case of having a pallet located somewhere in the environment with a free operating face, i.e., it can be eventually be picked up by an AGV as an autonomous forklift. It consists of 340 samples. 

\item \textit{NoPallet} class represents the case in which no pallet is present in the environment, or there is, but the operating face is too cluttered to allow an AGV such as an autonomous forklift to pick up the pallet. It consists of 225 samples. 
\end{enumerate}

\noindent This manual labeling step has been performed with the help of an online ROS visualization tool, \textit{RViz}\footnote{\url{http://wiki.ros.org/rviz}}. An operator checked the screen of the PC while the sensor was being moved, marking frames where a pallet with a free operating face was present in the sensor's FOV. 

Afterwards, any range data frame $R_i$ can also be represented as a set $S_i$ of polar coordinates, and consequently converted to Cartesian coordinates using \eqref{eq:2} and \eqref{eq:3}. 

\begin{equation}\label{eq:2}
s_i = \lbrace(r_0, \phi_0), \ldots, (r_j, \phi_j), \ldots, (r_{M-1}, \phi_{M-1})\rbrace.
\end{equation}	

\begin{equation}\label{eq:3}
\left\{
\begin{array}{l}
x_j = r_j\cos(\phi_j),\\
y_j = r_j\sin(\phi_j).
\end{array}
\right.
\;
\end{equation}

\begin{figure*}[p]
\centering
\vspace{-0.8in}
\begin{overpic}[width=0.33\textwidth,height=2.1in]{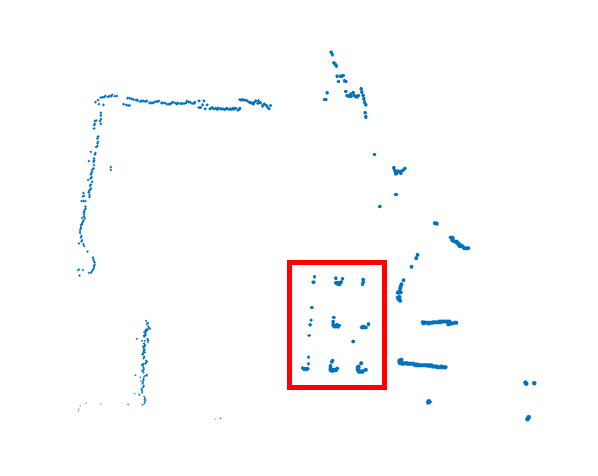}
\put (5,85) {\textit{Pallet}}
\end{overpic}
\begin{overpic}[width=0.33\textwidth,tics=10]{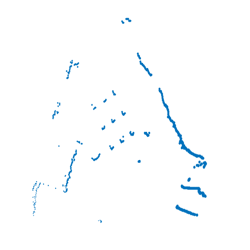}
\end{overpic}
\begin{overpic}[width=0.33\textwidth,tics=10]{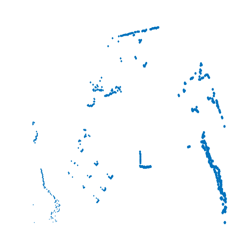}
\end{overpic}
\begin{overpic}[width=0.33\textwidth,tics=10]{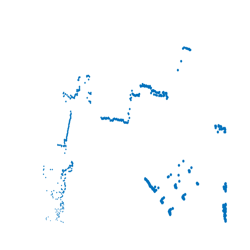}
\end{overpic}
\begin{overpic}[width=0.33\textwidth,tics=10]{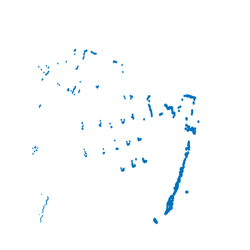}
\end{overpic}
\begin{overpic}[width=0.33\textwidth,tics=10]{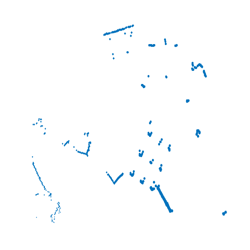}
\end{overpic}
\begin{overpic}[width=0.325\textwidth,height=2.1in]{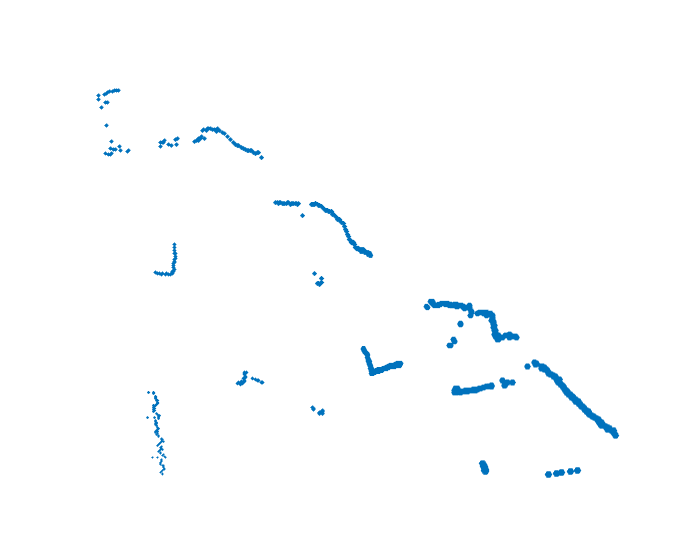}
\put (5,85) {\textit{NoPallet}}
\end{overpic}
\begin{overpic}[width=0.325\textwidth,height=2.1in]{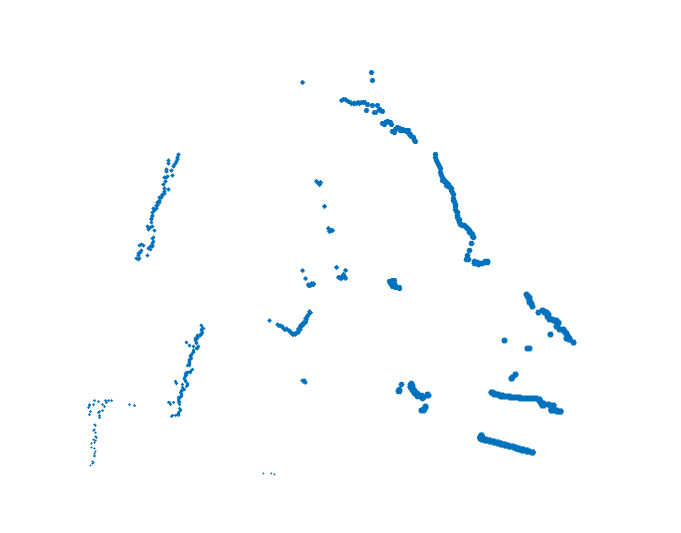}
\end{overpic}
\begin{overpic}[width=0.325\textwidth,height=2.1in]{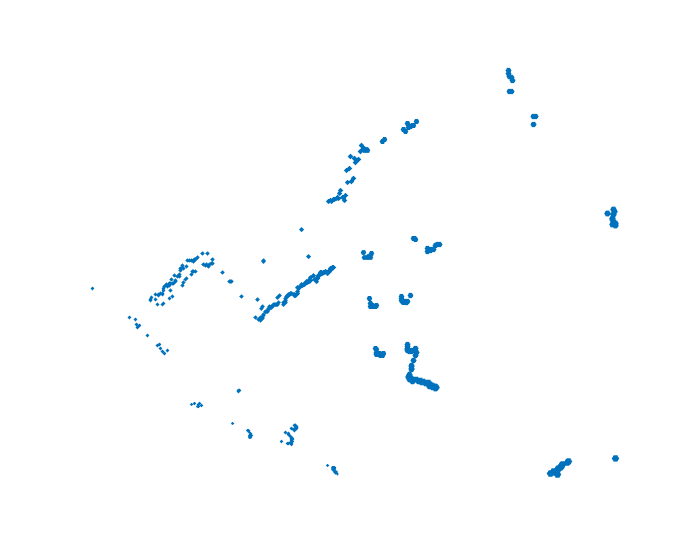}
\end{overpic}
\begin{overpic}[width=0.325\textwidth,height=2.1in]{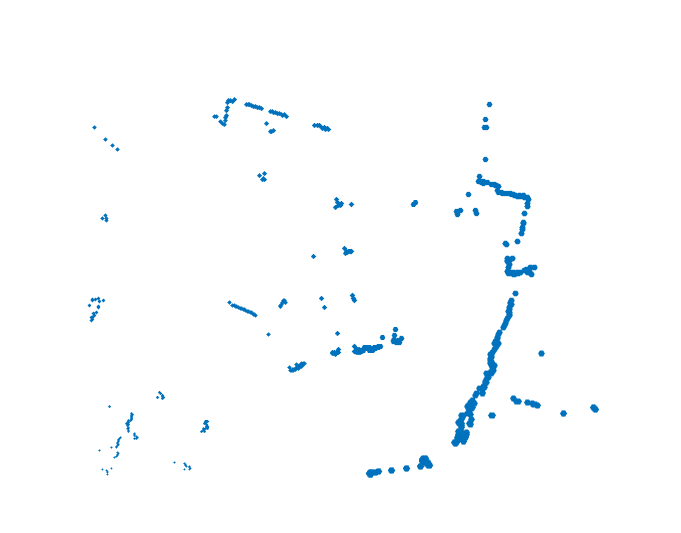}
\end{overpic}
\begin{overpic}[width=0.325\textwidth,height=2.1in]{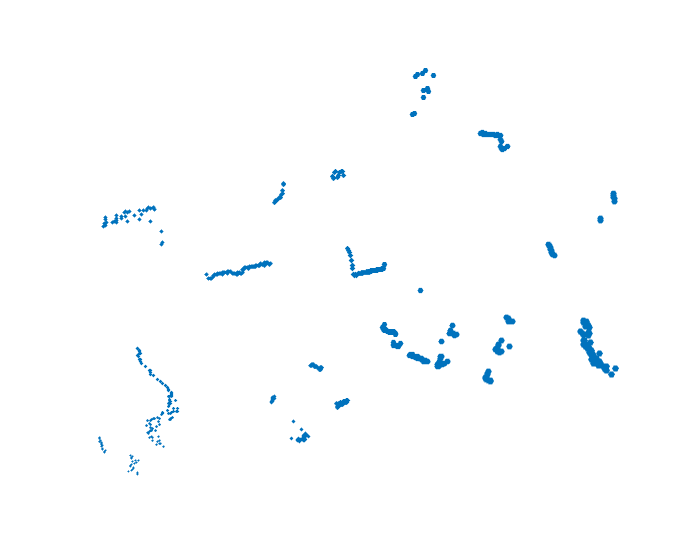}
\end{overpic}
\begin{overpic}[width=0.325\textwidth,height=2.1in]{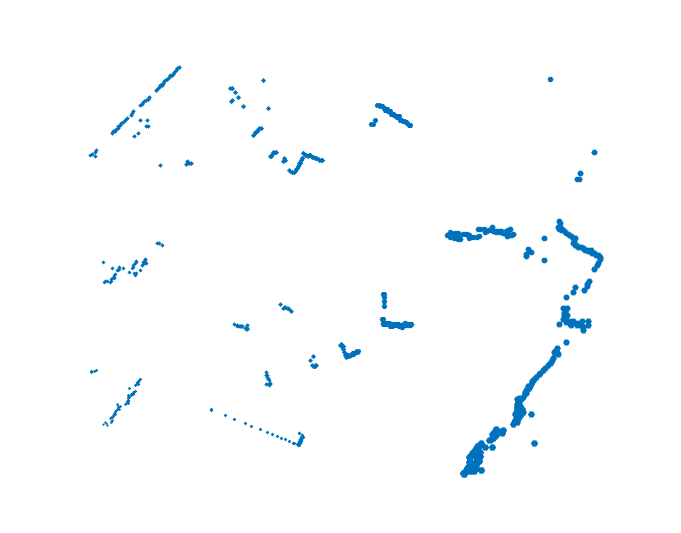}
\end{overpic}
\caption{The dataset of real-world 2D scans represented in Cartesian coordinates: the first two rows are related to the case where a pallet is present in the environment and the operating face is free, whilst the last two rows represent samples of the dataset when no pallet is present or the operating face is not accessible by an autonomous forklift. The red box in the first image represent an example of region of interest, i.e., the part of the image actually where the pallet is located.}	
\label{fig:pallet_examples}
\end{figure*}

\noindent This results in a binary 2D image of the operating area's floor plan, which is then resized to \unit[$250\times250$]{px}. An example of the resulting images is given in Fig. \ref{fig:pallet_examples}. 

Of course, these images are labeled with the same class as the originating frame. In partcular, images belonging to the \textit{Pallet} class come with the respective pallet ROI expressed as its upper-left and lower-right vertices (i.e., ($x_{min}, y_{min}$) and ($x_{max}, y_{max}$)), as well as a companion \unit[$250\times250$]{px} image containing the pallet only. Such ROIs are the results of the Region Proposal Network we employed in the related research paper \cite{mohamed2018detection}. The resulting ROIs have been manually labelled to indicate whether they present a pallet or another object. A selection of ROIs not including a pallet is also included in the dataset repository.

We will not further delve here into the details of our specific solution to the problem of pallet localization and tracking, which we present instead in the related research paper \cite{mohamed2018detection}. We just point out that the data was indeed employed for pallet localization and tracking and that the proposed architecture was tested using $4$ additional continous trajectories, which are also made available on the dataset repository. In particular, localization was performed using the aforementioned Region Proposal Network, cascaded with a Faster Recurrent Convolutional Neural Network classifier that took as input the set of manually labelled ROIs \cite{ren2015faster}. On the other hand, tracking was performed using a Kalman Filter \cite{cuevas2005kalman}. The filter was also used to implement a Sequantial Classification procedure, i.e., accepting a ROI as an actual pallet was deferred till it was detected and tracked for a predefined amount of time, eventually reaching a sufficient confidence threshold.

Finally, note that the dataset can be used for multi-pallet detection, but that was not part of our data collection experiment. Indeed, in the related research paper \cite{mohamed2018detection} we ran a preliminary study on the subject by generating artificial data. We want to stress that given that the EUR-Pallet is an official standard with strict tolerances, differences between any two pallets are not perceivable by the the sensor, due to its characteristics and margin of error. This leads to two major consequences:

\begin{itemize}
    \item It is not possible with this sensor and with the dataset we provide to univocally identify a pallet, yet it is possible to distinguish them from each other if appropriate tracking techniques are put in place, like we did in the related research paper \cite{mohamed2018detection}.
    \item Artificial 2D images including two or more pallets in every image are easy to generate. This can be achieved by taking an original image and adding the pallet ROI from another image, possibly changing position, orientation, and/or adding noise, and consequently deleting any reading in the original image that would now be occluded by the new pallet. Such artificial images are not provided here, but can easily be generated with the provided materials and tools. Nevertheless, future work on our related research paper will include real world multi-pallet testing, thus an extended dataset will be made available too.
\end{itemize}

\begin{table*}[ht!]
\vspace{0.1in}
\small\addtolength{\tabcolsep}{-4.2pt}
\hspace{0cm}\begin{tabular}{|c|c}
\hline
Index  & \multicolumn{1}{c|}{Range data $R_i$}\\
\hline\hline
\multicolumn{2}{|c|}{Frame \#$1$ ($i=1$)}\\
\hline\hline
0   &  \multicolumn{1}{c|}{3.11}\\
1   &  3.11\\
2   &  3.00 \\
\vdots  & \vdots \\
100 &  2.26 \\
101 &  2.28 \\
\vdots  & \vdots \\
$j$  & ${}^{1}r_{j}$ \\
\vdots  & \vdots \\
757 &  1.51\\
758 &  4.05 \\
759 &  4.08 \\
760 &  4.08 \\
\multicolumn{2}{c}{\scriptsize \hspace{-0.9cm} END of Frame \#$1$}\\
\end{tabular}
\hspace{-0.43in}\raisebox{-0.57in}{
\small\addtolength{\tabcolsep}{3pt}
\begin{tabular}{|c|c}
\hline
\multicolumn{2}{|c|}{Frame \#$2$ ($i=2$)}\\
\hline\hline
0   &  \multicolumn{1}{c|}{3.11}\\
1   & 3.11\\
2   & 3.11 \\
\vdots  & \vdots \\
100 &  2.26 \\
101 &  2.28 \\
\vdots  & \vdots \\
$j$  & ${}^{2}r_{j}$ \\
\vdots  & \vdots \\
757 &  1.51\\
758 &   4.08 \\
759 &   4.06 \\
760 &   4.08 \\
\multicolumn{2}{l}{\scriptsize END of Frame \#$2$}\\
\end{tabular}
}
\hspace{-0.45in}\raisebox{-1.03in}{
\small\addtolength{\tabcolsep}{3pt}
\begin{tabular}{|c|c}
\hline
\multicolumn{2}{|c|}{Frame \#$3$ ($i=3$)}\\
\hline\hline
0   &  \multicolumn{1}{c|}{3.13}\\
1   &  3.11\\
2   &  3.13 \\
\vdots  & \vdots \\
100 &  2.26 \\
101 &  2.28 \\
\vdots  & \vdots \\
$j$  & ${}^{3}r_{j}$ \\
\vdots  & \vdots \\
757 &  1.51\\
758 &   4.08 \\
759 &   4.08 \\
760 &   4.05 \\
\multicolumn{2}{l}{\scriptsize END of Frame \#$3$}\\
\end{tabular}}
\hspace{-0.35in}\raisebox{-1.5in}{
\small\addtolength{\tabcolsep}{3pt}
\begin{tabular}{|c|c|}
\hline
\multicolumn{2}{|c|}{Frame \#$4$ ($i=4$)}\\
\hline\hline
0   &  3.13\\
1   &  3.11\\
2   &  3.00 \\
\vdots   & \vdots \\
100 &  2.23 \\
101 &  2.28 \\
\vdots  & \vdots \\
$j$  & ${}^{4}r_{j}$ \\
\vdots  & \vdots \\
757 &  1.48\\
758 &   4.08 \\
759 &   4.08 \\
760 &   4.08 \\
\multicolumn{2}{|c|}{\scriptsize END of Frame \#$4$}\\
\hline\hline
\end{tabular}}
\hspace{-0.7in}\raisebox{0.5in}{$\xLongrightarrow[\text{over $4$ frames}]{\text{Taking \textit{average}}}$}\hspace{0.1in}
\small\addtolength{\tabcolsep}{-0.2pt}
\begin{tabular}{|c|c|}
\hline
Index  & \multicolumn{1}{c|}{Range data $R_i$}\\
\hline\hline
0   &  3.12\\
1   &  3.11\\
2   &  3.06 \\
\vdots  & \vdots \\
100 &  2.252 \\
101 &  2.28 \\
\vdots  & \vdots \\
$j$  & $\frac{{}^{1}r_{j}+\cdots+{}^{4}r_{j}}{4}$ \\
\vdots  & \vdots \\
757 &  1.50\\
758 &   4.07 \\
759 &   4.075 \\
760 &   4.07 \\
\hline\hline
\end{tabular}
\caption{An example of the raw range data provided by the laser rangefinder. As soon as the data is visualized using the standard ROS package \textit{rviz}, four sequential frames are stored in a text file. Then, the \textit{average} can be calculated in order to perform the detection and tracking of the pallet using machine learning techniques.}
\label{TAB:label}
\end{table*}

\subsection{Dataset inspection}
The dateset is completely contained in the \textit{AllData} folder of the provided git repository. The folder is structured as follows:

\begin{itemize}
    \item The \textit{Class1} and \textit{Class2} folders correspond to \textit{Pallet} and \textit{NoPallet} classes, respectively. They include $565$ raw laser rangefinder scans in \textit{.txt} format in total, $340$ for the former class and $225$ for the latter.
    \item \textit{DataSet565.mat} is a file containing the whole dataset as a $761 \times 565$ MATLAB matrix.
    \item \textit{PalletImages} folder containing all the \unit[$250\times250$]{px} images in various formats, divided by class and eventually accompained by the relative pallet's ROI. In particular, the files \textit{PalletGrayImages.zip} and \textit{RGBImages.tar.gz} contains the images in \textit{.jpg} and \textit{.png} format, respectively.
    \item \textit{TrajectoryDataset} folder contains $4$ additional continous trajectories that we used to test the architecture presented in our related research paper \cite{mohamed2018detection}. The trajectories are provided in \textit{.mat} format.
\end{itemize}

\section*{Acknowledgements}
The work by I. S. Mohamed was supported by a scholarship from the ERASMUS+ European Master on Advanced Robotics Plus (EMARO+) programme. The authors would like to thank M.Eng. Yusha Kareem for his helping in data collection process.

\section*{Conflict of interest}
The authors declare that they have no conflict of interest relevant to this article.

\section*{Transparency document. Supplementary material}
Transparency data associated with this article can be found in the online version at \url{https://github.com/EmaroLab/PDT}.

\bibliographystyle{elsarticle-num}
\section*{References}
\bibliography{mybibfile}
\end{document}